\documentclass[runningheads]{llncs}

\usepackage{graphicx,xspace,amsmath,booktabs,multicol,caption,subcaption,wrapfig,hyperref,bbding}
\hypersetup{hidelinks,backref=true,pagebackref=true,hyperindex=true,colorlinks=true,breaklinks=true,urlcolor= blue,bookmarks=true,bookmarksopen=false,pdftitle={`Title},pdfauthor={Author}}

\begin{document}

\title{V-Zen: Efficient GUI Understanding and Precise Grounding With A Novel Multimodal LLM}
\titlerunning{V-Zen}

\author{Abdur Rahman (\Envelope)\orcidID{0000-0002-9547-2435} \and
Rajat Chawla \and
Muskaan Kumar \and
Arkajit Datta \and
Adarsh Jha \and
Mukunda NS \and
Ishaan Bhola}

\authorrunning{A. Rahman et al.}

\institute{SuperAGI Research\\
\email{abdur75648, rcrajatchawla\{@gmail.com\}, muskaan, arkajit, adarsh, mukunda, ishaan \{@superagi.com\}}}

\maketitle 

\begin{abstract}
    In the rapidly evolving landscape of AI research and application, Multimodal Large Language Models (MLLMs) have emerged as a transformative force, adept at interpreting and integrating information from diverse modalities such as text, images, and Graphical User Interfaces (GUIs). Despite these advancements, the nuanced interaction and understanding of GUIs pose a significant challenge, limiting the potential of existing models to enhance automation levels. To bridge this gap, this paper presents V-Zen, an innovative Multimodal Large Language Model (MLLM) meticulously crafted to revolutionise the domain of GUI understanding and grounding. Equipped with dual-resolution image encoders, V-Zen establishes new benchmarks in efficient grounding and next-action prediction, thereby laying the groundwork for self-operating computer systems. Complementing V-Zen is the GUIDE dataset, an extensive collection of real-world GUI elements and task-based sequences, serving as a catalyst for specialised fine-tuning. The successful integration of V-Zen and GUIDE marks the dawn of a new era in multimodal AI research, opening the door to intelligent, autonomous computing experiences. This paper extends an invitation to the research community to join this exciting journey, shaping the future of GUI automation. In the spirit of open science, our code, data, and model will be made publicly available, paving the way for multimodal dialogue scenarios with intricate and precise interactions. Repo Link: \href{https://github.com/abdur75648/V-Zen}{github.com/abdur75648/V-Zen}

\end{abstract}

\section{Introduction}
\label{section:intro_section}

Introduction In the vibrant and ever-evolving field of artificial intelligence, Multimodal Large Language Models (MLLMs)\cite{mmllms1,mmllms2} have emerged as a transformative force, bridging the gap between diverse data representations and their comprehension. These models, adept at integrating information from multiple modalities such as text and images, have significantly expanded the scope of research and practical applications. A critical area of focus within this domain is the automation of tasks involving Graphical User Interfaces (GUIs)\cite{Koh2024VisualWebArenaEM}. The automation of these tasks holds immense potential for enhancing efficiency and productivity across a wide range of applications.

\begin{figure}[t]
	\includegraphics[width=\textwidth]{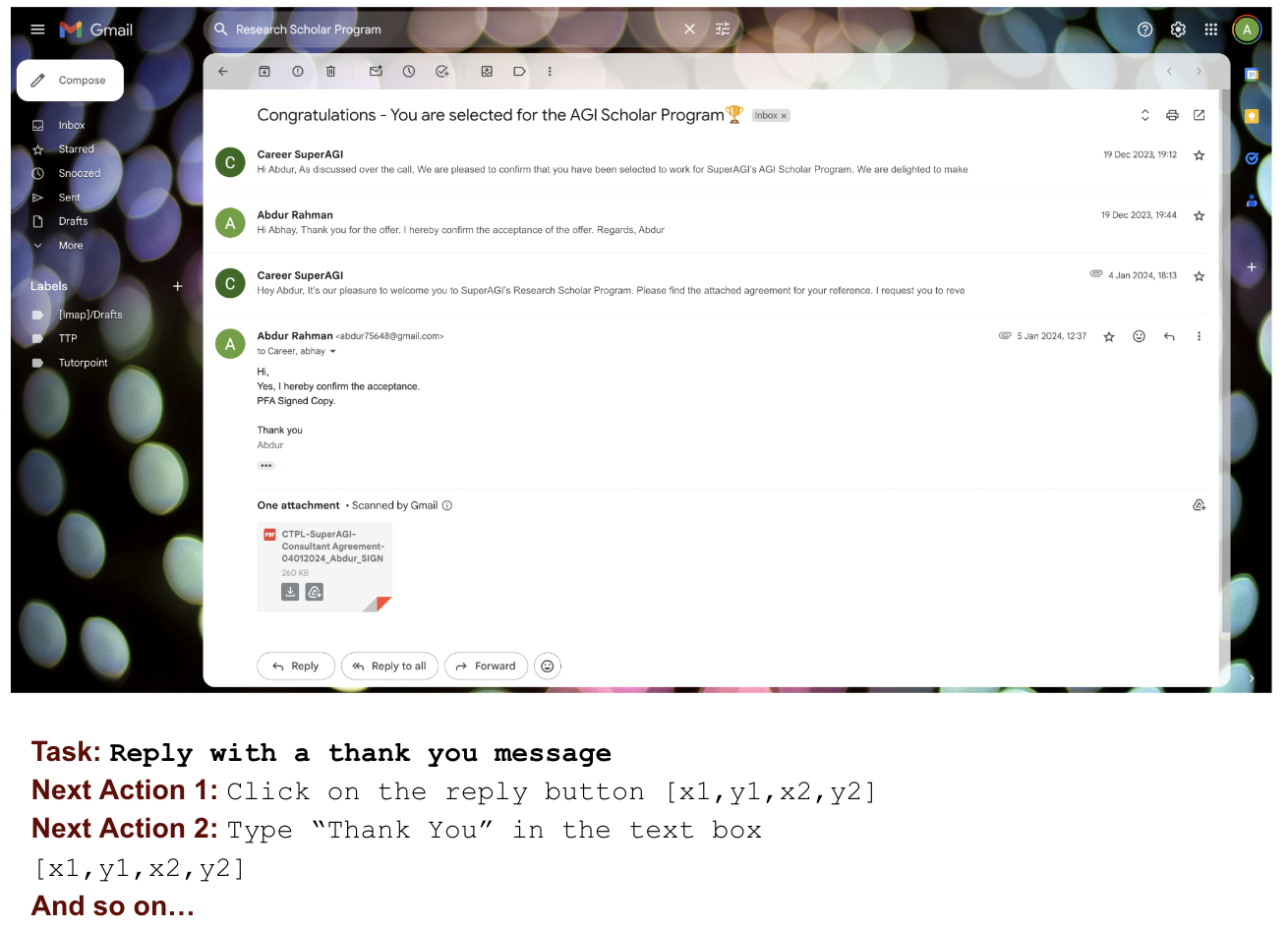}
	\caption{A Sample Case of GUI Automation Difficulty. In order to build intelligent systems capable of interacting seamlessly with various applications, identifying relevant UI components is crucial. As shown in this Gmail example, specifying tasks and their logical continuations requires a precise understanding of underlying GUI structures, predicting the next action, and precisely performing the grounding task. Our approach addresses these challenges effectively.}
	\label{fig:label1}
\end{figure}

\begin{figure}[t]
	\includegraphics[width=\linewidth]{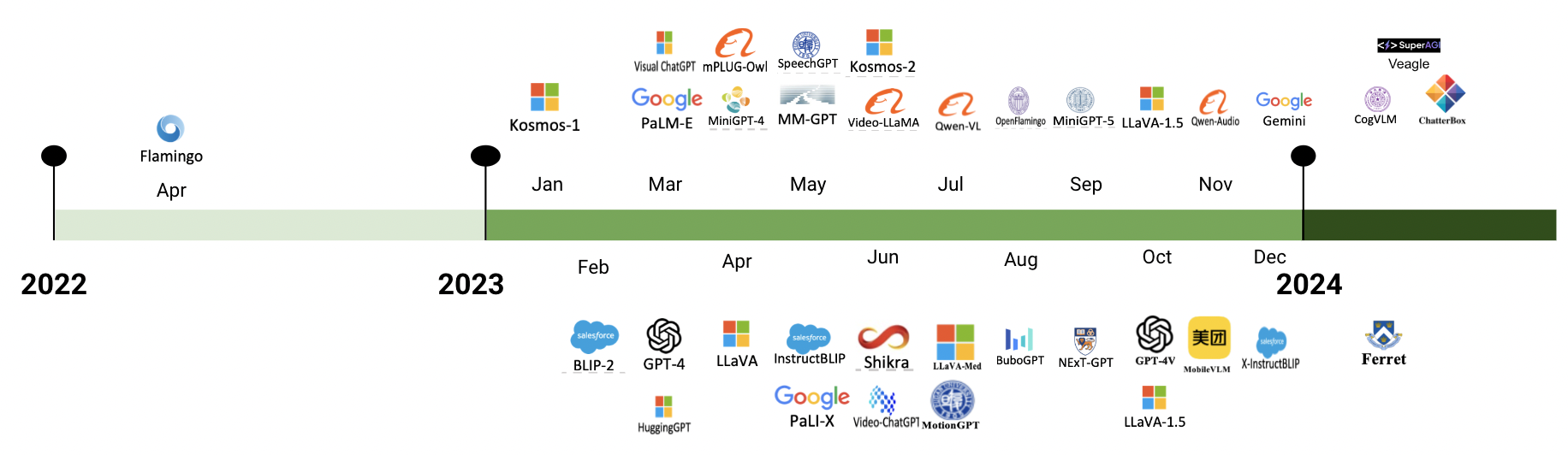}
	\caption{A timeline of SOTA MLLMs}
	\label{fig:label2}
\end{figure}

However, a significant portion of existing models and benchmarks in this field have been primarily centred on text-based tasks. This approach overlooks the vast potential of multimodal agents that can effectively process and integrate visual information for problem resolution. The main thrust of our research is the application of these models, with a particular emphasis on the concept of grounding, especially in the context of GUI images. Grounding, in the realm of MLLMs, refers to the process of associating words or phrases in a language with corresponding entities in other modalities. For instance, in a text-image pair, the term “apple” would be grounded in the image of an apple. This capability of MLLMs to efficiently and precisely perform grounding is particularly crucial for automating GUI tasks \cite{grounding1,grounding2}.

However, grounding in MLLMs presents a unique set of challenges. A primary concern is the alignment of modalities, i.e., ensuring the model accurately correlates entities across different modalities. Several multimodal LLMs have recently addressed this issue by employing projection layers to convert one embedding to another. Despite these advancements, the coordinates of the bounding boxes provided by these models in the form of LLM text responses often lack precision. This issue becomes particularly pronounced when dealing with GUIs, where the accuracy of object detection and localization is critical. Existing LLMs rely on textual descriptions of visual content or the HTML context of web pages, but essential details like icons, images, diagrams, and spatial relationships may be overlooked or misrepresented during conversion to text embeddings \cite{gui_automation1,gui_automation2}. Many GUIs do not offer direct textual interfaces for automation, highlighting the need for a multimodal LLM that can directly process visual GUI signals. The precision in detecting and interacting with GUI elements is of paramount importance in this context. The ability to accurately identify and interact with GUI elements not only enhances the functionality of these agents but also significantly augments their utility in real-world applications. The primary objective of this research is to push the boundaries of multimodal agent-based GUI task automation by developing a Multimodal Large Language Model (MLLM) that can effectively navigate, understand, and interact with GUI elements with high precision.

Our proposed model, V-Zen, is specifically designed to address these challenges. V-Zen is not just another MLLM but a sophisticated GUI Agent that can accurately process image-text inputs, interpret natural language instructions, precisely identify GUI elements, and execute actions on websites to accomplish user-defined objectives. V-Zen integrates a visual grounding module that harnesses the capabilities of the DINO detector, equipping it to effectively handle multimodal grounding tasks. In addition to the text response by LLM, the coordinates of grounding are provided separately by the grounding module, replacing a typical object detection module, thereby ensuring precise coordinates. The model’s performance is further augmented by a High Cross-Resolution Module (HRCM), which enables the model to process high-resolution features and comprehend text within images.
In conjunction with the development of the novel model, we have also created a dataset for this task named GUIDE (Graphical User Interface Data for Execution)\cite{guide_paper}, a cutting-edge benchmark dataset that includes bounding box annotations and textual descriptions with chain of thought collected across various GUI platforms. GUIDE aids in advancing agentive research, ultimately leading to the development of more agile, responsive, and human-like agents across a multitude of fields.

Our key contributions in this paper are:
\begin{enumerate}
	\item We propose V-Zen, a novel GUI Agent that leverages the power of MLLMs for efficient GUI understanding and task prediction, forming a self-operating system for various GUI tasks.
        
	\item We introduce a visual grounding module that leverages the DINO detector’s capabilities, enabling it to handle multimodal grounding tasks effectively.
	
	\item We design a unique architecture that processes an input image in parallel at two different resolutions, allowing for efficient GUI understanding and task prediction.
	
	\item We curate and publicly release GUIDE, a state-of-the-art benchmark dataset for executing tasks on diverse GUI platforms.
	
\end{enumerate}

In addition to our key contributions outlined above, we conduct a thorough comparative analysis of state-of-the-art (SOTA) Grounding MLLM models under similar experimental setups. We also examine the contributions of individual modules in our model towards accuracy as an ablation study (Table 3). Finally, we discuss the remaining limitations and potential avenues for future research in the field.
The remainder of the paper is structured as follows: Section 2 offers a comprehensive review of related work in the field of MLLMs and grounding. Section 3 delineates the architecture of our proposed model, V-Zen. Section 4 introduces the GUIDE dataset and its construction. Section 5 discusses the experiments conducted and the results obtained. Finally, Section 6 concludes the paper and outlines future work. This research aims to contribute significantly to the field of AI, pushing the boundaries of what is possible in GUI automation.

\section{Related Work}
\label{section:related_work_section}

The field of Natural Language Processing (NLP) has witnessed a significant transformation with the advent of Large Language Models (LLMs\cite{naveed2024comprehensivellms,minaee2024large}). GPT-3 \cite{gpt3}, one of the pioneering LLMs, marked a milestone by significantly scaling up the model size and training data size, showcasing exceptional performance in numerous NLP tasks and setting a trend for subsequent advancements in the field. Several models such as GPTs \cite{gpt_summary}, PaLM \cite{palm_paper}, BLOOM \cite{bloom_paper}, and LLaMA \cite{llama_paper}, have since emerged, each pushing the boundaries of LLMs. These models have demonstrated remarkable abilities in learning from in-context examples, reasoning, following instructions, and operating over long-context sequences. Recent endeavours in the field have concentrated on refining LLMs to better align with human instructions and feedback, with models like InstructGPT \cite{instruct_gpt}, ChatGPT \cite{gpt3}, and GPT4 \cite{gpt4} standing out as exemplars in this regard.

In the context of building web agents, these LLMs have been leveraged extensively. However, they are primarily text-based and lack the capability to handle images or other modalities. This limitation has led to the development of Multimodal Large Language Models (MLLMs). MLLMs extend the capabilities of LLMs to understand and integrate information from multiple modalities, such as vision and audio \cite{mmllms1}. In the context of GUI automation, our primary focus is on MLLMs, where the input modalities include text and image, and the output is a corresponding text response. The architecture and functioning of MLLMs can vary, but they generally follow a similar pattern: An encoder for each data modality generates the embeddings for data of that modality, an embedding layer aligns embeddings of different modalities into the same multimodal embedding space, and then a LLM generates text responses. Models like Flamingo \cite{alayrac2022flamingo}, Kosmos-1 \cite{huang2023kosmos1}, BLIP-2 \cite{li2023blip2}, and PaLM-E \cite{palm_paper} exemplify this approach. Over time, the inherent reasoning and decision-making capabilities of MLLMs have improved, enabling them for more intricate tasks like image retrieval, image generation, and visual question answering.

The application of MLLMs in grounding tasks has been a significant area of research. Works such as Kosmos-2 \cite{peng2023kosmos2} and Shikra \cite{chen2023shikra} have enabled MLLMs to perform fine-grained image comprehension and open-world grounding. Additional works in this direction include GPT4ROI \cite{zhang2023gpt4roi}, PVIT \cite{chen2023pvit}, BuboGPT \cite{zhao2023bubogpt}, VisionLLM \cite{wang2023visionllm}, Ferret \cite{you2023ferret}, Veagle \cite{veagle_paper} and CogVLM \cite{wang2024cogvlm}. While these works improve the grounding capabilities of the model through architectural improvements or training strategy improvements, they all have a few limitations, which our work aims to address. Firstly, they produce bounding boxes in the form of pure text output, which, even if it points to the correct object, is not highly accurate. This is particularly relevant for GUI automation tasks, where there are several small elements in GUIs that need to be accurately grounded for some tasks. Secondly, most of them commonly use a 224 × 224 resolution image input, which makes the tiny icons and texts in GUI screenshots difficult to recognize.

Our proposed model, V-Zen, addresses these challenges by introducing a novel architecture for efficient GUI understanding and precise grounding. For accurate grounding of GUI elements, we introduce a separate grounding module on top of the LLM in the style of an open-set object detection model, and we also enable a high-resolution 1120 × 1120 image input through a cross-attention branch inspired by CogVLM. Additionally, we meticulously curate an extensive instruction-tuning dataset for executing tasks on diverse GUI platforms and fine-tune our model on it. As a result, V-Zen exhibits superior performance compared to previous works when it comes to executing tasks on diverse GUI platforms.

\section{Proposed Architecture}
\label{section:architecture_section}

\begin{figure}[t]
	\includegraphics[width=\textwidth]{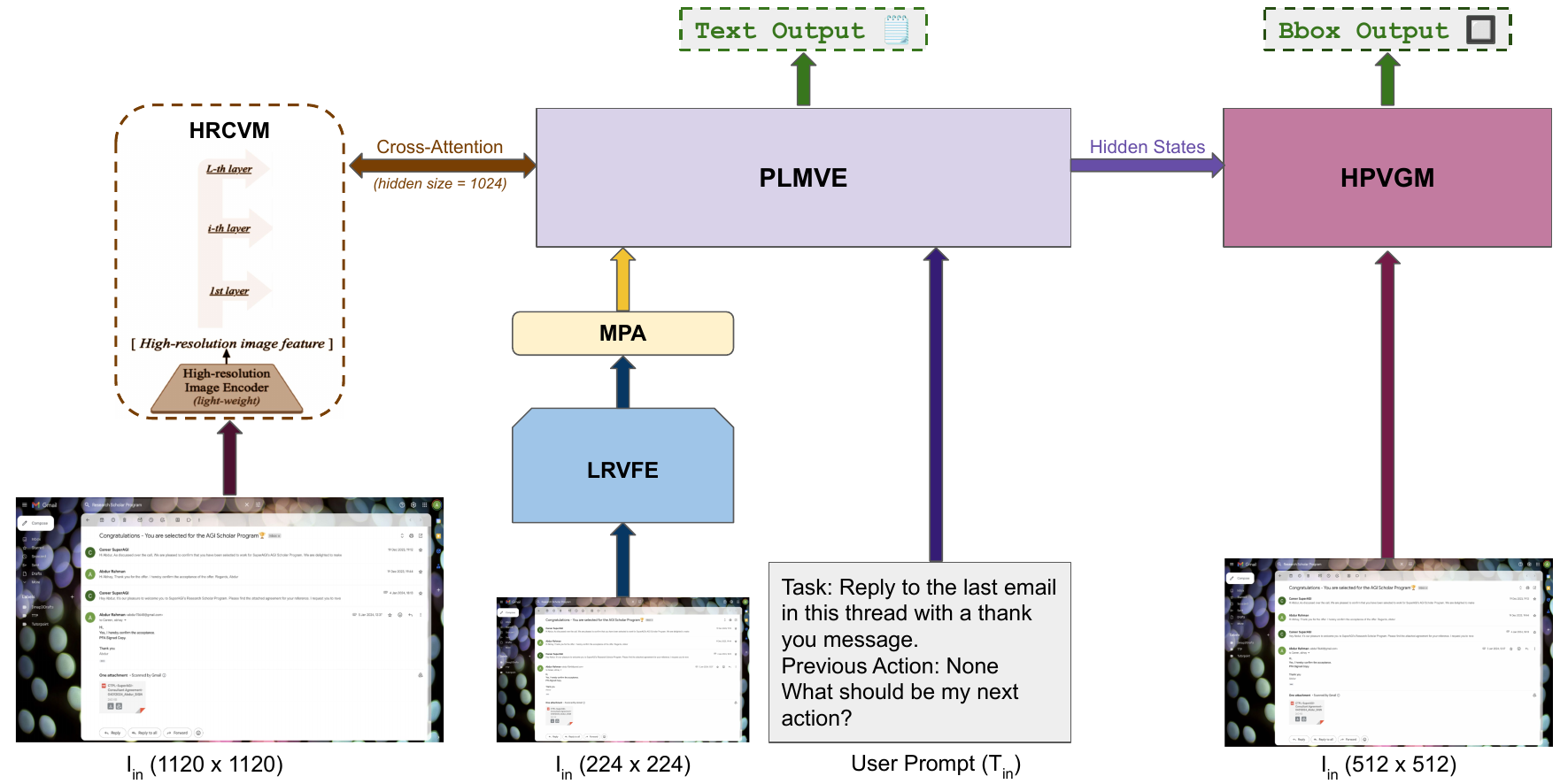}
	\caption{Proposed Architecture Of V-Zen.}
	\label{fig:label3}
\end{figure}

The architecture of V-Zen, our proposed multimodal Large Language Model (LLM), is a sophisticated ensemble of interconnected components meticulously designed for efficient GUI understanding and precise grounding. The architecture is composed of five major modules: Low-Resolution Visual Feature Extractor (\textit{LRVFE}), Multimodal Projection Adapter (\textit{MPA}), Pretrained Language Model with Visual Expert (\textit{PLMVE}), High-Resolution Cross Visual Module (\textit{HRCVM}), and High-Precision Visual Grounding Module (\textit{HPVGM}).

\subsection{Low-Resolution Visual Feature Extractor}
\label{label4}

The journey of input through the architecture commences with the LRVFE, a low-resolution encoder (EVA-2-CLIP \cite{evaclip_paper,clip_paper}) that processes the input image at a resolution of 224x224. This module is responsible for extracting meaningful features from the image, which are then used for further processing. Given an input image $I_{in}$ and text prompt $T_{in}$, the LRVFE generates low-resolution image features as:

\begin{center}
$F_{LR} = LRVFE(I_{in})$
\end{center}

\subsection{Multimodal Projection Adapter}
\label{label5}

The features extracted by the LRVFE are transformed by the MPA into a format that is suitable for processing by the LLM backbone of our architecture \cite{llava_paper}. The MPA plays a pivotal role in aligning the modalities, ensuring that the image features match the input format of the LLM. The transformation can be represented as:

\begin{center}
$F_T = MPA(F_{LR})$
\end{center}

, where $F_T$ are the transformed features.
\subsection{Pretrained Language Model with Visual Expert}
\label{label6}

The PLMVE, which adopts Vicuna-7B \cite{vicuna2023} as the base language model, is tasked with generating text outputs based on the processed image features and any textual input provided. Given an input $X_{in}^{(i)}$ to the $ith$ attention layer of the PLMVE, it’s split into $X_{img}^{(i)}$ and $X_{txt}^{(i)}$. Then, $Q_{img}^{(i)}$ is obtained as:

\begin{center}
$Q_{img}^{(i)} = VEL(X_{img}^{(i)})$
\end{center}

, and $Q_{txt}^{(i)}$ is obtained as:

\begin{center}
$Q_{txt}^{(i)} = OLL(X_{txt}^{(i)})$
\end{center}

The overall output can be represented as:

\begin{center}
\end{center}

This can be overall represented as:

\begin{center}
\end{center}

, where VEL represents the Visual Expert Layers, OLL represents the original LLM Layers, and MHSVE represents the process of multi-head self-attention with the visual expert.

\subsection{High-Resolution Cross Visual Module}
\label{label7}

The HRCVM, inspired by CogAgent \cite{hong2023cogagent}, is designed for higher-resolution input, accepting images of size 1120 × 1120 pixels. It employs a smaller EVA2-CLIP vision encoder and cross-attention of a small hidden size to fuse high-resolution image features with every layer of the PLMVE. This can be represented as

\begin{center}
$X_{hi} = HRCVM(I_{HR})$
\end{center}

, where $I_{HR}$ is the high-resolution input image, and $X_{hi}$ is the high-resolution output of the HRCVM. Each layer’s attention procedure with the residual connection can be formulated as

\begin{center}
$X_{out}^{(i)} = MHSVE(X_{in}^{(i)}) + X_{in}^{(i)}$
\end{center}

And then final output features with residual connection can be formulated as:

\begin{center}
$Y_{out}^{(i)} = MHCA(X_{out}^{(i)}, X_{hi}) + X_{out}^{(i)}$
\end{center}

, where MHCA represents multi-head cross-attention.

\subsection{High-Precision Visual Grounding Module}
\label{label8}

The HPVGM takes the hidden states extracted from the PLMVE and uses them to perform precise grounding tasks \cite{chatterbox,contextdet}. Unlike typical MLLM modules that provide grounding bounding boxes as part of the LLM’s text output, our HPVGM outputs bounding box coordinates separately, ensuring precision. The module follows an enhanced DETR \cite{carion2020DETR} object detector named DINO \cite{groundingDINO}. PLMVE’s last hidden state is used as the query of visual grounding to query the multi-scale feature set for visual grounding, denoted as $q_{llm\_gnd}$.
The multi-scale feature set, denoted as $f{ms\_img}$, is obtained using a Swin Transformer-based backbone. It takes $q_{llm\_gnd}$ and $f{ms\_img}$ and produces the bounding boxes for precise grounding. This way, the HPVGM module can precisely ground the GUI elements based on the processed image and text features.

In conclusion, the architecture of V-Zen, our proposed multimodal Large Language Model (LLM), represents a sophisticated orchestration of several interconnected components. Each module within this architecture is meticulously designed and plays a pivotal role in achieving the overarching goal of efficient GUI understanding and precise grounding. The design of these modules and their intricate interconnections is a testament to the detailed planning and innovative thinking that has gone into the development of V-Zen. This complex yet efficient assembly of components not only enhances the functionality of the system but also significantly augments its utility in real-world applications. The architecture, therefore, stands as a robust framework that pushes the boundaries of what is possible in GUI automation, ultimately contributing significantly to the field of artificial intelligence.

    

    

\begin{table}[t]
\centering
\setlength{\tabcolsep}{10pt}
\begin{tabular*}{\textwidth}{@{\extracolsep{\fill}}lp{5cm}}
    \toprule[1.5pt]
    \textbf{Method} & \textbf{Accuracy} \\ \hline
    
    Base Model with LRVFE and Vicuna & 87.5 \\
    *+HRCVM & 89.6 \\
    *+Grounding DINO & 90.3  \\
    *+Projection Layer & 92.9  \\
    *+Mistral LLM & 93.2  \\
    \bottomrule[1.5pt] \\
\end{tabular*}
\caption{Ablation Study wrt Next Task Prediction.}
\label{tab:label10}
\vspace{-2em}
\end{table}

\begin{table}[t]
\centering
\setlength{\tabcolsep}{10pt}
\begin{tabular*}{\textwidth}{@{\extracolsep{\fill}}lp{5cm}}
    \toprule[1.5pt]
    \textbf{Method} & \textbf{Accuracy} \\ \hline
    
    Base Model with LRVFE and Vicuna & 74.5 \\
    *+HRCVM & 76.2 \\
    *+Grounding DINO & 89.1  \\
    *+Projection Layer & 89.7  \\
    *+Mistral LLM & 89.7  \\
    \bottomrule[1.5pt] \\
\end{tabular*}
\caption{Ablation Study wrt Grounding.}
\label{tab:label10}
\vspace{-2em}
\end{table}


\begin{table}[h]
\centering
\resizebox{8.5 cm}{!}{
\begin{tabular}{llllllll}
\hline
\vspace{0.5em}
 & \textbf{Next Task Prediction} & \textbf{Grounding} \\ \hline
GPT-4V & 94 & 28 \\
Gemini-Pro & 92 & 21 \\
Chatter-Box & 91.3 & 87.9 \\
CogAgent & 92.4 & 86.3 \\
V-Zen & \textbf{93.2} & \textbf{89.7} \\
\hline
\end{tabular}
}
\vspace{2em}
\caption{Performance of the proposed model.}
\label{tab:accuracy}
\end{table}

\section{Experiments and Results}
\label{section:experiments}

\subsection{Training Procedure}

Following the CogAgent \cite{hong2023cogagent} pre-training strategy, the model undergoes a two-stage training procedure consisting of pre-training and specialised fine-tuning (SFT). During pre-training, the focus lies on enhancing the model's ability to grasp high-resolution images and adapt them for GUI applications by emphasising text recognition, visual grounding, and understanding GUI imagery. Various public datasets serve as pre-training resources, covering synthetic renderings, academic documents, and optical character recognition (OCR) images. After completing the pre-training stage, SFT uses the GUIDE dataset, a specially curated collection of real-world GUI elements and task-based sequences. Through fine-tuning, V-Zen learns from complex workflows, action histories, and negative samples, gaining proficiency in making accurate inferences and performing pertinent actions on previously unencountered GUIs. Training benefits from NVIDIA's 8*A100 platform and utilises the DeepSpeed library for optimal speed while applying the Adam optimiser, a learning rate of 0.00001, a batch size of 8, and a gradient accumulation step of 1 to maintain steady learning progression.

\begin{figure}[t]
	\includegraphics[width=\textwidth]{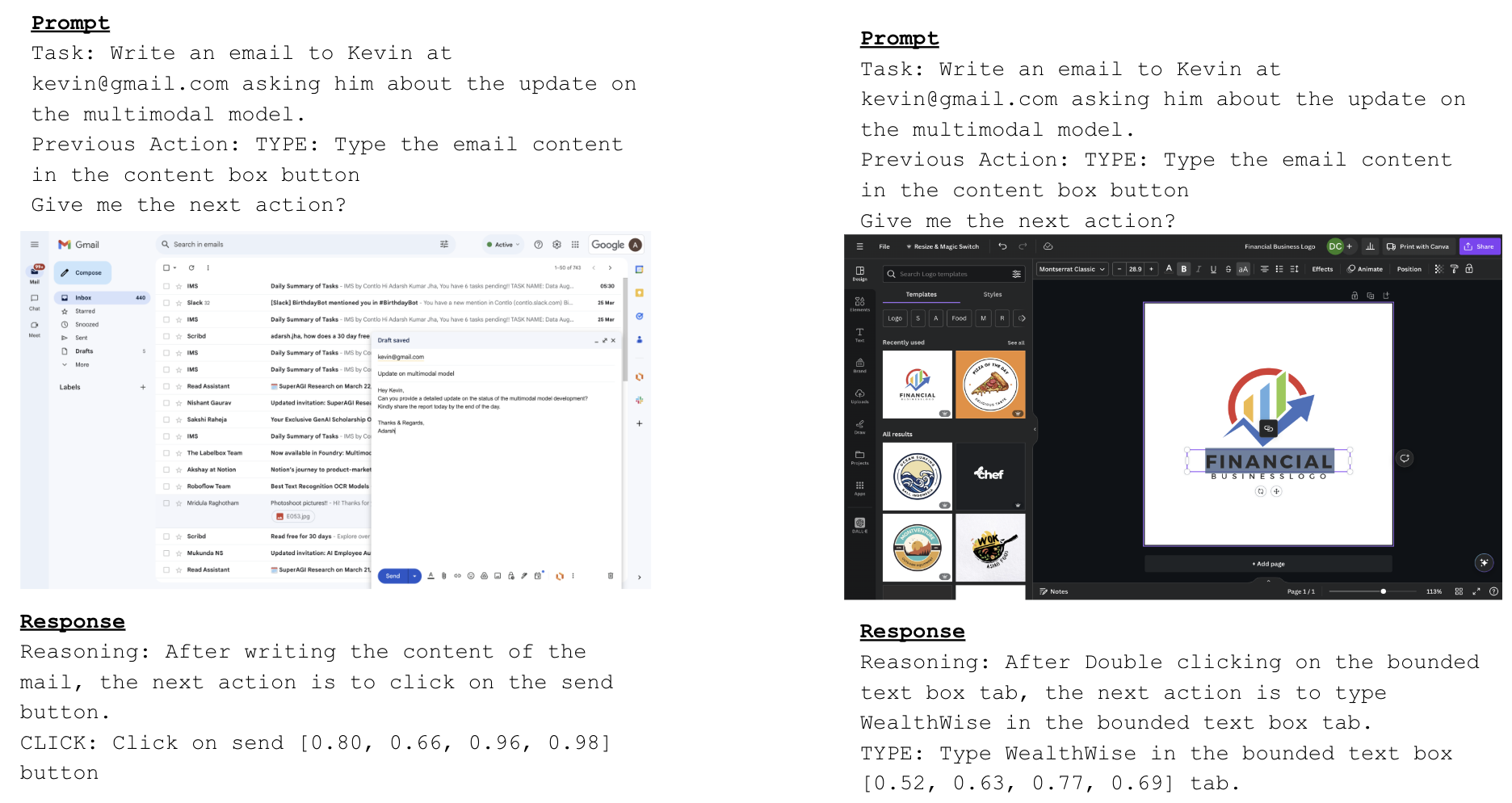}
	\caption{Some samples of the GUIDE dataset: Notice how the next action is predicted along with the bounding box locations, demonstrating the dataset's utility in guiding Multimodal Large Language Models for GUI automation tasks.}
	\label{fig:label3}
\end{figure}

\subsection{GUIDE Dataset}
The GUIDE (Graphical User Interface Data for Execution) \cite{guide_paper} dataset is a large-scale, meticulously curated dataset developed specifically to enhance the applications of Multimodal Large Language Models (MLLMs), with a particular focus on Robotic Process Automation (RPA) use cases. The dataset, which comprises 124,000 data points, authentically represents user interactions within various GUI environments and covers a diverse array of fields, online platforms, and activities. It includes data from popular GUI platforms such as Apollo.io, Contlo, Gmail, Google Calendar, and Canva. Each data entry in GUIDE consists of an image, a task description, the last action taken, and the next action to be performed, along with grounding information indicating where the action needs to be executed. Furthermore, the dataset incorporates a Chain of Thought (CoT), preserving historical records of earlier actions and promoting contextual reasoning during model operation. The dataset was collected using the authors’ in-house advanced annotation tool, NEXTAG (Next Action Grounding and Annotation Tool), and adapted for multiple operating systems, browsers, and display types. It was collated by multiple annotators to capture the variation of design and the way a person uses a website. GUIDE supports investigations into cross-interface automated tasks and encourages the development of multi-platform LLMs for practical applications in automation and natural language understanding. In essence, GUIDE is about predicting the next task on a given GUI image and performing the corresponding grounding task for correctly interacting with GUI elements like boxes, buttons, icons, etc., across a diverse range of platforms.

\subsection{Results And Discussion}
In this section, we delve into the empirical evaluation of our proposed model, V-Zen, and its performance on the GUIDE dataset. The evaluation focuses on two pivotal tasks: Next Task Prediction and Grounding. For the Next Task Prediction, we assess the model’s ability to predict the next action accurately. Specifically, we compare the predicted action with the ground-truth action in terms of semantic meaning. To measure accuracy, we consider an action prediction correct if it aligns with the intended task progression. For grounding, we focus on bounding box localization accuracy. The F1 score, commonly used in object detection tasks, serves as our primary evaluation metric for grounding accuracy. We juxtapose the performance of V-Zen with other state-of-the-art models, namely CogAgent, GPT-4V, Chatterbox, and Gemini-Pro, under similar experimental conditions to ensure a fair comparison.
As delineated in Table 1, V-Zen exhibits superior performance in the Next Task Prediction task, achieving an accuracy of 93.2\%. This metric is indicative of V-Zen's proficiency in accurately predicting the subsequent task in a GUI environment, thereby demonstrating its potential in real-world applications. In the context of the Grounding task, V-Zen continues to outperform the other models, as evidenced in Table 3. With a next-task prediction accuracy of 93.2\% and grounding accuracy of 89.7\%, V-Zen demonstrates its capability to precisely ground GUI elements, a critical aspect in GUI automation tasks.

These empirical results underscore the efficacy of V-Zen in both tasks, thereby attesting to its robustness and versatility. The success of V-Zen can be attributed to its innovative architecture, which seamlessly integrates low-resolution and high-resolution visual modules, a multimodal projection adapter, and a high-precision grounding module. This intricate design enables V-Zen to effectively process and integrate visual and textual information, thereby enhancing its GUI understanding and grounding capabilities. Furthermore, the use of the GUIDE dataset for specialised fine-tuning has significantly bolstered V-Zen's proficiency in handling real-world GUI elements and task-based sequences. The GUIDE dataset, with its diverse array of GUI environments and task-based sequences, provides a rich resource for training, thereby enabling V-Zen to learn from complex workflows, action histories, and negative samples. In conclusion, the experimental results substantiate the effectiveness of V-Zen in automating GUI tasks, thereby setting a new benchmark in the realm of multimodal large language models for GUI automation. The results presented herein provide a promising direction for future research in this domain. Future work will focus on further enhancing the performance of V-Zen and expanding its applicability to a wider range of GUI platforms.

\begin{figure}[t]
	\includegraphics[width=\textwidth]{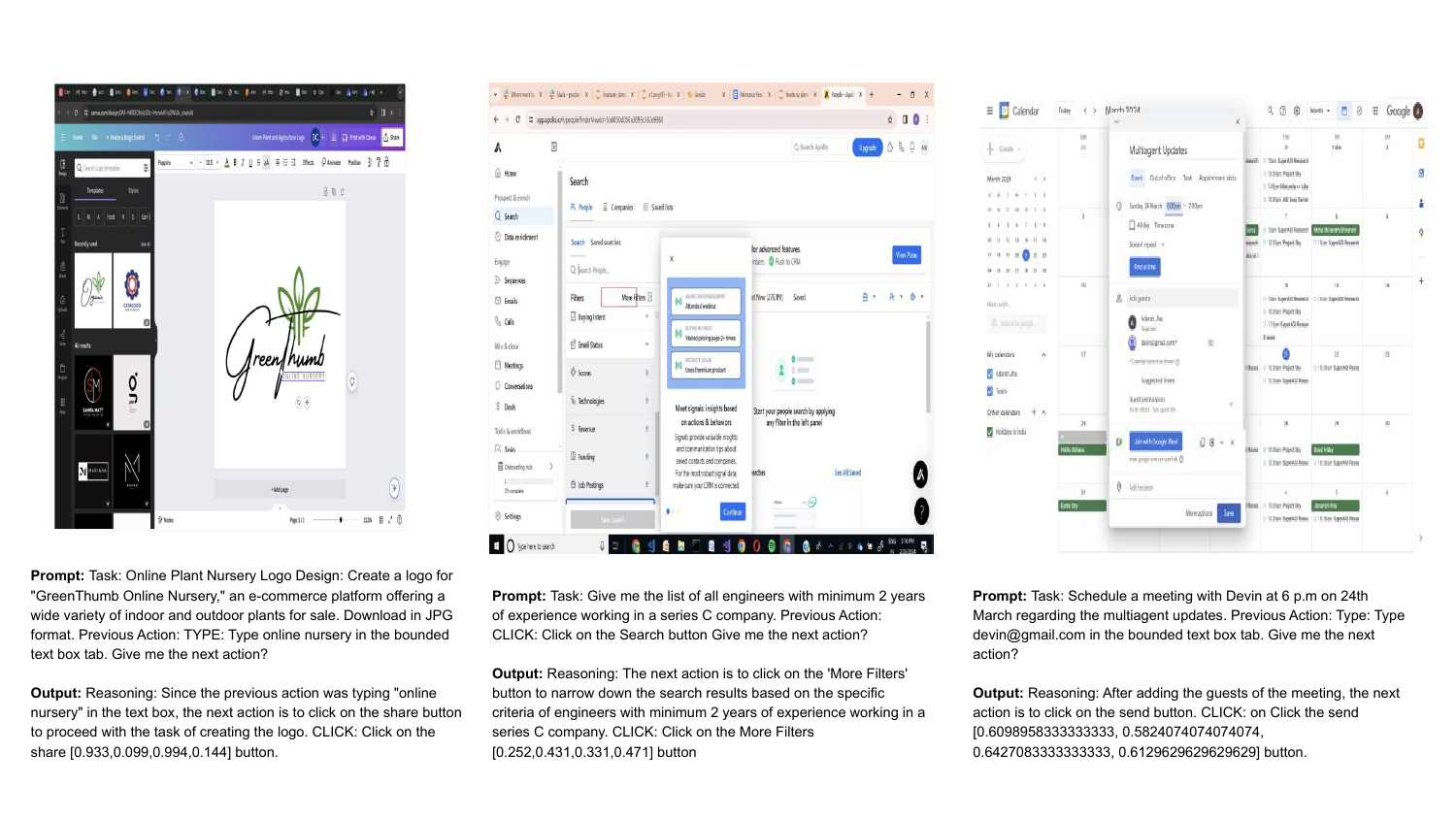}
	\caption{Qualitative Results on GUIDE Samples Using V-Zen. Demonstrates the effectiveness of our developed model in predicting the next actions and bounding box locations for achieving a given task.}
	\label{fig:label3}
\end{figure}

\section{Conclusion}
\label{section:conclusion}

In conclusion, this research paper has presented V-Zen, a groundbreaking Multimodal Large Language Model (MLLM) specifically engineered to revolutionise the realm of Graphical User Interface (GUI) understanding and grounding. V-Zen, with its innovative dual-resolution encoding mechanism and dedicated grounding module, has successfully transcended traditional limitations in GUI interaction and interpretation, thereby marking a significant advancement in GUI-centric AI solutions. Our rigorous evaluations have unequivocally demonstrated V-Zen's superior performance over competing models in next-action prediction and grounding tasks, thereby establishing it as a pioneering force in the domain of self-operating computer systems.

Simultaneously, we have introduced GUIDE, a state-of-the-art benchmark dataset meticulously compiled to catalyze advancements in MLLMs, with a particular emphasis on Robotic Process Automation (RPA) applications. GUIDE, with its comprehensive collection of GUI grounding-oriented dialogues and realistic spatial relationship quandaries, serves as a powerful catalyst propelling the field towards innovative breakthroughs in multimodal AI modeling.

The introduction of V-Zen and GUIDE marks a significant advancement in AI, setting the stage for future developments in this dynamic field. Our contributions aim to inspire future MLLMs, providing them with the tools needed to master GUI automation. We foresee continuous refinement of V-Zen, accommodating a wider range of GUI platforms and real-life complexities. Concurrently, we expect GUIDE to evolve, embracing complex and diverse scenarios to meet the growing demands of the field. Ultimately, we aspire to foster an ecosystem where AI can effectively tackle real-world problems, delivering value and contributing to societal betterment. The successful synthesis of V-Zen and GUIDE opens a new chapter in multimodal AI research, unlocking possibilities for intelligent, autonomous computing experiences. We invite fellow researchers to join us in shaping this exciting frontier, anticipating a future where AI not only enhances human capabilities but also enriches human experiences.

\bibliographystyle{splncs04}
\bibliography{refs}

\end{document}